# DataSist: A Python-based library for easy data analysis, visualization and modeling

Rising .O. Odegua[1], Festus .O. Ikpotokin[2]
Ambrose Alli University, Ekpoma, Edo State, Nigeria
[1]risingodegua@gmail.com, [2]festus@aau.edu

**Abstract.** A large amount of data is produced every second from modern information systems such as mobile devices, the world wide web, Internet of Things, social media, and so on. Analysis and mining of these massive data require a lot of advanced tools and techniques. Therefore, big data analytics and mining is currently an active and trending area of research because of the enormous benefits businesses and organizations derive from it. Numerous tools like pandas, numpy, STATA, SPSS, have been created to help analyze and mine these huge outburst of data and some have become so popular and widely used in the field. This paper presents a new python-based library, DataSist, which offers high level, intuitive and easy to use functions, and methods that can help data scientists/analysts to quickly analyze, mine and visualize big data sets. The objectives of this project are: to design a python library to aid data analysis process by abstracting low level syntax and to increase productivity of data scientist by making them focus on what to do rather than how to do it. This project shows that data analysis can be automated and much faster when we abstract certain functions, and will serve as an important tool in the workflow of data scientists.

## 1    Introduction

According to ScienceDaily, over 90% of the data in the world was generated in approximately two years [2]. This shows that big data has really come to stay and therefore new research and studies must be carried out in order to fully understand the massive data. This means there has to be a paradigmatic shift from past theories, technologies, techniques and approaches in data mining and analysis in order to fully harness the gold resident in these data [1]. Big data as noted in [3], has been coined to represent this outburst of massive data that cannot fit into traditional database management tools or data processing applications. These data are available in three different formats such as structured, semi-structured and unstructured and the sizes are in scales of terabytes and petabytes. Formally, big data is categorized into dimensions in terms of the 3Vs (see Figure 1), which are referred to as volume, velocity and variety [8] . Each of the three Vs make traditional operation on big data complicated. For instance, the velocity i.e speed at which the data comes has become so fast that traditional data analytical tools can not handle them properly and may breakdown when used. Also the increase in volume has made the extraction, storage and preprocessing of data more difficult and challenging as both analytical algorithms and system must be scalable in other to cope and these were not built into traditional systems from the onset. Lastly, the ever changing variety of data and its numerous source of integration makes the storage and analysis of data difficult.

The growth of big data has been exponential, and from the perspective of information and communication technology, it holds the key to better and robust products for businesses and organizations. This outburst as we have stated earlier comes with its own difficulties as regard analysis and mining, and this has been a major hindrance in the massive adoption of big data analytics by many businesses. The major problems here is the lack of effective communication between database systems with the necessary tools such as data mining and statistical tools. These challenges arise when we generally want to discover useful trends and information in data for practical application.

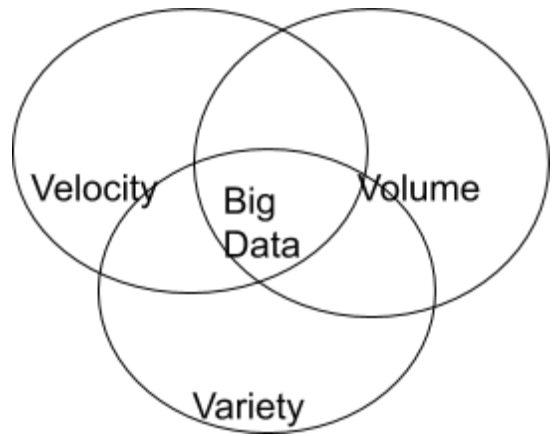

**Fig. 1.** The three Vs of big data

With big data comes big responsibilities. This is true and is the reason why most accumulated data in industries such as health care, retail, wholesale, scientific research, web based applications, among others, are dormant and underutilized. In order to fix these problems, we have to implement and understand the numerous ways to analyze big data. Similarly, it has been observed that many analytical techniques built to solve these problems are either too low level to learn easily, too specific to a task, or do not scale to big data. This necessitated the creation or when possible redesign of existing tools for the sole purpose of solving this problem.

Numerous shortcomings have been identified in most of the Python tools created to solve the challenges mentioned above. Top among them is lengthy lines of code for simple tasks, too low level for users, and no support for popular repetitive functions used during analysis. To solve these challenges, we created DataSist, a Python-based library for easy analysis, data mining and visualization. The functionalities of DataSist is encapsulated in several Python modules which handles different aspects of a data scientist's workflow. DataSist aims to reduce the difficulties in mining data by providing easy, high level, abstracted syntax that are easy to write and remember, and increase productivity and efficiency of the users, by making them focus on what to do rather than how to do it.

## 2    Literature Review

Big data is large; hence the information resident in it must be mined, captured, aggregated and communicated by data scientists/analysts. In order to fully and effectively carry out these tasks, data analysts are expected to have a specific kind of knowledge and to leverage on powerful big data analytics tools. Although there exists lots of tools for big data mining; they are broadly categorized into three groups namely, statistical tools, programming languages and visualization tools [10].

Many statistical tools like SPSS, STATA, Statistica, among others, are popular for big data analysis but the choice of which to use vary among data analysts as the usage dictates the choice of tool. The SPSS (Statistical Package for the Social Sciences) is a widely used tool in the field, though it was originally built for the social sciences [23], SPSS software is now used by market researchers, data miners [24], survey companies, health researchers and others.

STATA is another popular statistical tool used mainly by analysts. One major problem with STATA is that unlike SPSS, it is more difficult to use by people from non-statistical background. STATA is a general purpose tool and is used mainly in the field of economics, sociology, medicine and computer science [25].

Programming languages like Python and R are also famous for data analysis and mining. The high level, dynamic and interactive nature of these languages combined with the abundance of scientific libraries make them a preferred choice for analytical tasks [17]. While R is still the most popular language for traditional data analytical tasks, the usage of Python language has been on the increase since the early 2000s, in both industry applications and research [21]. These has led to the development of the Python ecosystem of libraries, some of which are Numpy, Pandas, Scipy and IPython. These are briefly explained below:

Numpy (Numerical Python), offers fast and efficient interface for working with high multi-dimensional arrays and matrices. Numpy is a base data structure and fundamental package in Python and as such numerous libraries are also built on top of it.

Pandas is a popular package built on top of Numpy that is used for data manipulation and analysis [19]. It offers efficient data structures and operations for manipulating data usually in tabular forms. Some of the features available in Pandas are DataFrame objects for data manipulation, tools for reading and writing data between memory, hierarchical axis indexing functions to work with high dimensional data, time series functionality, data filtration and data alignment for handling missing data. The library is highly optimized as core modules are written in C and Cython.

Scipy is also another popular library in Python scientific ecosystem. It is a collection of packages for efficient computation of linear algebra, matrix, special functions and numerous statistical functions.

IPython is an interactive computing and development environment used mainly by data scientists and analyst for rapid prototyping and exploration of data [16]. IPython is web based usually in the form of a notebook, that offers rich GUI consoles for inline plotting, exploratory analysis, inline coding and markdown text functionalities.

The R programming language is versatile and extremely popular open source language used by data scientists and statisticians [15]. R offers functional based syntax for performing numerous statistical functions and has powerful debugging facilities. It also offers high level and powerful graphical tools [13]. Some of the features that make R a popular choice for analysts is that it has a short and slim syntax, a long and robust list of packages for numerous analytical tasks, availability on numerous OS and platforms and numerous variants for loading and storing data. Some factors that affect adoption is the higher learning curve for people from non-statistical background [11].

Another important aspect of big data analysis is visualization. As a result, numerous tools and software have been built to aid effective data visualization. Most of the programming languages like Python and R have their own plotting packages some of which are R's ggplot, Python's matplotlib, seaborn, bokeh, plotly, JavaScript d3.js, and so on. There have also been massive development of GUI visualization tools, and some popular ones are Tableau, Power Bi, Qlikview, Spotfire and Google Data Studio [5].

**2.1 Data Analysis**. Data analysis is the process of inspecting, cleaning, transforming and modeling data with the purpose of discovering useful information, getting actionable insights, supporting decision making and informing conclusions. Data analysis has multiple approaches including diverse techniques which depends heavily on the problem. This means there is no single or fixed approach to conducting data analysis. In today's world, data analysis plays a crucial role in making decisions and helping businesses operate more efficiently. Some data analysis techniques include Data mining, Business intelligence, Descriptive statistics. These are briefly explained below:

*Data Mining.* Data mining is a particular type of data analysis that focuses on predictive modeling rather than descriptive modeling. Data mining is an interdisciplinary subfield of computer science and statistics with the ultimate goal of extracting patterns and knowledge from large datasets. It uses numerous methods like machine learning, statistics and database systems [14].

*Business Intelligence.* Business Intelligence (BI) deals primarily with data concerned with businesses. It is a combination of technologies and tools used by businesses for analysis of business information. It also helps inform decision making, identify, develop and create strategic business opportunities, and gives businesses a competitive market advantage [12].

In statistical applications, data analysis may be classified into descriptive statistics, confirmatory data analysis, predictive analysis and exploratory data analysis.

*Descriptive Statistics.* Descriptive statistics provide detailed summaries about observations or sample of data. These statistics could be quantitative, summary statistics like mean, mode, medians, percentiles, max and min etc. or visual, such as graphs and plots. Descriptive statistics form a basic and gives the analyst an intuition into the underlying data set. In business, descriptive statistics helps in summary of many types of data. For example, Marketers and sales personnel may use buyers historical spending and buying patterns by performing simple descriptive analytics on the data in order to make better product decisions. Descriptive statistics may be divided into Univariate, Bivariate and Multivariate analysis. These are briefly explained below:

*Univariate analysis.* Univariate analysis is one of the simplest ways for describing data. The prefix "Uni" means "one", meaning the analysis deals with one feature at a time. This means that when performing Univariate analysis, we do not consider causal relationships among features but instead the main purpose is to describe a single feature. The most popular descriptive statistics found in univariate analysis include central tendency (mean, mode and median) and dispersion (range, variance, maximum, minimum, quartiles and standard deviations. Using graphs and charts, there are several types of univariate analysis we can perform, some of which are Bar Charts, Histograms, Frequency Polygons and Pie Charts.

*Bivariate analysis*. Bivariate analysis is the analysis of two features compared side by side, in order to find possible relationships between them [4]. The result of bivariate analysis can be used to answer the question of whether a feature "X" depends on another feature "Y", whether there is a causal or linear dependence among these features and whether one can help predict another. Some popular types of bivariate analysis include scatter plots, regression analysis and correlation coefficients.

*Multivariate analysis*. Multivariate analysis is the analysis of three or more features and the relationship among them. It is more complex than both univariate and bivariate analysis. This type of analysis is mostly performed using special tools and softwares like Pandas, SAS, SPSS etc., as working with three or more data features manually is infeasible. Multivariate analysis is mostly preferred when the data set under consideration is diverse, and each feature or relation among features is important [6]. Multivariate analysis has applications in numerous domains some which are dimensionality reduction, Clustering, Variable selection, Classification analysis, discrimination analysis and Latent structure discovery.

## 2.2 Data Processing

Raw data is almost always useless and must be transformed into a usable form. For example, processing may include converting data into structured form (tabular) as most Python analytical tools can work well with this format, cleaning and removal of outliers or useless features, filling or removal of missing values and standardization/normalization of the data values.

## 2.3 Data Cleaning

Data cleaning is often a part of the data processing phase. Data cleaning comes after data has been processed and organized. In this stage, data received may contain duplicates, redundant features, errors, or be incomplete. This means the data needs to be cleaned for use. Some of the common task done in this stage may include data deduplication, data normalization, feature segmentation, record matching etc.

## 2.4 Data Visualization

Data visualization is the graphical representation of data. It includes all the processes and techniques involved in the communication of data in pictorial form that aids easy understanding and communication. This communication is achieved through the use of systematic mapping of graphical points to data values. In order to communicate information present in data, data visualization makes use of statistical graphics, charts, plots and info-graphics, etc. As remarked in [9], the main goal of data visualization is to communicate information clearly and effectively through graphical means. To convey ideas from big data effectively, both aesthetics, interpretability and functionality must go side by side. The major aim of data visualization is the provision of insights into complex and diverse data set by showing the key-aspects in a more intuitive way.

## 3 Implementation and Architecture

DataSist has been implemented using the Python programming language, and its design is currently centered around five modules (feature_engineering, modeling, structdata, timeseries and visualization). The feature_engineering module contains functions to handle tasks such as cleaning, filling of missing values, aggregating and counting features in a dataset. The modeling module contains functions used for machine learning modeling, structdata module handles all tasks relating to structured data in tabular form, timeseries module handles temporal features such as date-time, timestamps and finally, the visualization module holds numerous functions that aids easy visualization of features in a dataset (see Fig. 2).

DataSist has been created for Interactive use in Jupyter Notebook environments and has been integrated into the workflow of data analyst in Python. It builds on a lot of existing libraries like the efficient DataFrame and Series from Pandas, uses Numpy's fast matrices and array functions and heavily depends on Seaborn and Matplotlib for visualization.

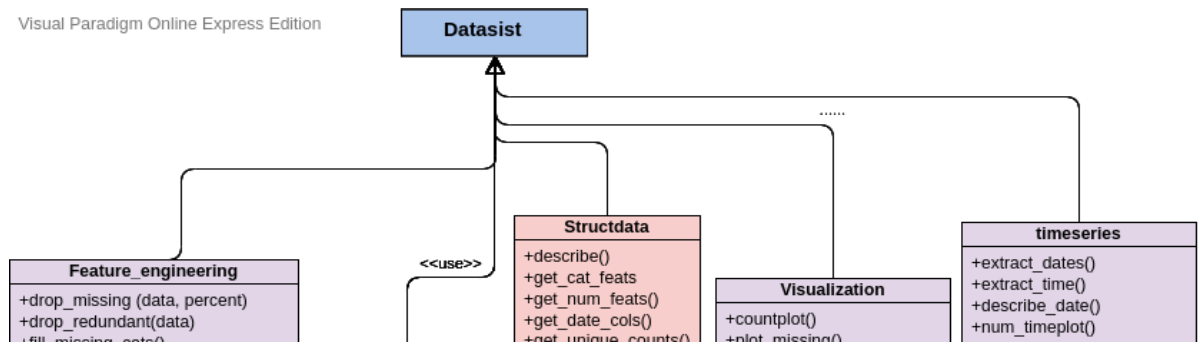

**Fig. 2.** Class diagram showing modules and their member functions.

### 3.1 Usage

To demonstrate an end-to-end use of the DataSist library, we obtain a dataset from the competitive data science platform Zindi [26]. The dataset is part of a predictive machine learning challenge hosted by Xente [27] an e-payments, e-commerce and financial service company in Uganda. The dataset contains samples of approximately 140,000 transactions between 15th November 2018 and 15$^{th}$ March 2019, and the task is to build a machine learning model from the data to detect if a transaction is fraudulent or not.

We perform this analysis in an interactive coding environment called Jupyter Notebook [17]. First, we import the necessary libraries to use including the DataSist library (see Fig. 3), then, we do a quick summary or description using the describe function in the structdata module of our library (see Fig. 4a, 4b and 4c). The describe function helps to give a detailed summary of the important attributes and characteristics of the dataset. Some of the descriptive statistics displayed by the describe function are:

1. First, last and random five instances of the dataset.
2. Shape and size of the dataset.
3. Data types found in the dataset.
4. List of Date features found in the dataset.
5. List of categorical and numerical features found in the dataset.
6. Statistical description of the numerical features in the dataset including count, mean, standard deviation, min and max.
7. Unique classes found in the categorical features.
8. Percentage of missing values found in the dataset.

After describing the dataset, we remove redundant features using the drop_redundant function. (see Fig. 5). These features are irrelevant in a mining task as they have low variance, and contain only a single class.

After the initial cleaning, we can do some visualizations to help us understand the dataset better. This can easily be done using functions available in the visualization module of our library. We start by doing visualization for the categorical features. Here, we may use functions such as countplot and catbox.

Countplot shows the unique classes in a categorical feature and their corresponding size (see Fig. 6). This helps us to know the most common class in a feature. Catplot on the other hand, makes a plot of all categorical features and separates them by a categorical target (see Fig. 7) . This is useful in classification modeling task, as it helps show which of the classes are important in the separation of the target. To visualize numerical features in the dataset, we can use the histogram or boxplot function in the visualization module. The histogram helps to show the univariate distribution of values in a feature (see Fig. 8), while the boxplot shows the distribution of values based on calculated quartiles such as median, 25th, 50th and 75th percentiles and they can help detect outliers.

Next, we can explore temporal features in the dataset using the timeseries module. The function num_timeplot available in this module can be used to plot all numeric features against a selected time feature. This can help us check for patterns and seasonality in a dataset (see Fig 9). The function extract_dates can be used to extract date information like hour of the day, minute of the day, day of the month, day of the year, day of the week, from a time feature easily (see Fig. 10).

Finally, we demonstrate the modeling process of this analysis using the model module. This module contains functions like train_classifier, get_classification_report and plot_feature_importance that helps us train and test a classification model. First, we import the necessary machine learning models, split the data into local train and validation set, and then create the model instances (see Fig. 11). We can display a detailed performance report of a trained model such as Lightgbm and RandomForest classifiers (see Fig. 12) and also plot the important features to a model (see Fig. 13).

```python
import pandas as pd
import numpy as np
import datasist as ds

train_data = pd.read_csv('training.csv')
test_data = pd.read_csv('test.csv')
```

**Fig. 3.** Importing Python libraries for data analysis

```python
ds.structdata.describe(train_data)
```

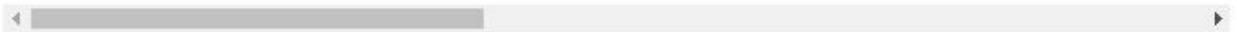

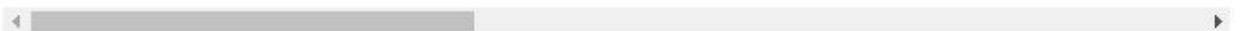

**Fig. 4a.** Output from using the describe function on the dataset

```
Column(s) {'TransactionStartTime'} should be in Datetime format. Use the
[to_date] function in datasist.feature_engineering to coonvert to Pandas D
atetime format

Numerical Features in Data set
['CountryCode', 'Amount', 'Value', 'PricingStrategy', 'FraudResult']
```

Statistical Description of Columns

|       | CountryCode | Amount        | Value         | PricingStrategy | FraudResult  |
|-------|-------------|---------------|---------------|-----------------|--------------|
| count | 95662.0     | 9.566200e+04  | 9.566200e+04  | 95662.000000    | 95662.000000 |
| mean  | 256.0       | 6.717846e+03  | 9.900584e+03  | 2.255974        | 0.002018     |
| std   | 0.0         | 1.233068e+05  | 1.231221e+05  | 0.732924        | 0.044872     |
| min   | 256.0       | -1.000000e+06 | 2.000000e+00  | 0.000000        | 0.000000     |
| 25%   | 256.0       | -5.000000e+01 | 2.750000e+02  | 2.000000        | 0.000000     |
| 50%   | 256.0       | 1.000000e+03  | 1.000000e+03  | 2.000000        | 0.000000     |
| 75%   | 256.0       | 2.800000e+03  | 5.000000e+03  | 2.000000        | 0.000000     |
| max   | 256.0       | 9.880000e+06  | 9.880000e+06  | 4.000000        | 1.000000     |

```
Data Types
Note: All Non-numerical features are identified as objects in pandas
```

|                      | Data Type |
|----------------------|-----------|
| TransactionId        | object    |
| BatchId              | object    |
| AccountId            | object    |
| SubscriptionId       | object    |
| CustomerId           | object    |
| CurrencyCode         | object    |
| CountryCode          | int64     |
| ProviderId           | object    |
| ProductId            | object    |
| ProductCategory      | object    |
| ChannelId            | object    |
| Amount               | float64   |
| Value                | int64     |
| TransactionStartTime | object    |
| PricingStrategy      | int64     |
| FraudResult          | int64     |

**Fig. 4b.** Output from using the describe function on the dataset

Missing Values in Data

| | features | missing_counts | missing_percent |
|---|---|---|---|
| 0 | TransactionId | 0 | 0.0 |
| 1 | BatchId | 0 | 0.0 |
| 2 | AccountId | 0 | 0.0 |
| 3 | SubscriptionId | 0 | 0.0 |
| 4 | CustomerId | 0 | 0.0 |
| 5 | CurrencyCode | 0 | 0.0 |
| 6 | CountryCode | 0 | 0.0 |
| 7 | ProviderId | 0 | 0.0 |
| 8 | ProductId | 0 | 0.0 |
| 9 | ProductCategory | 0 | 0.0 |
| 10 | ChannelId | 0 | 0.0 |
| 11 | Amount | 0 | 0.0 |
| 12 | Value | 0 | 0.0 |
| 13 | TransactionStartTime | 0 | 0.0 |
| 14 | PricingStrategy | 0 | 0.0 |
| 15 | FraudResult | 0 | 0.0 |

Unique class Count of Categorical features

| | Feature | Unique Count |
|---|---|---|
| 0 | TransactionId | 95662 |
| 1 | BatchId | 94809 |
| 2 | AccountId | 3633 |
| 3 | SubscriptionId | 3627 |
| 4 | CustomerId | 3742 |
| 5 | CurrencyCode | 1 |
| 6 | ProviderId | 6 |
| 7 | ProductId | 23 |
| 8 | ProductCategory | 9 |
| 9 | ChannelId | 4 |
| 10 | TransactionStartTime | 94556 |

**Fig. 4c.** Output from using the describe function on the dataset

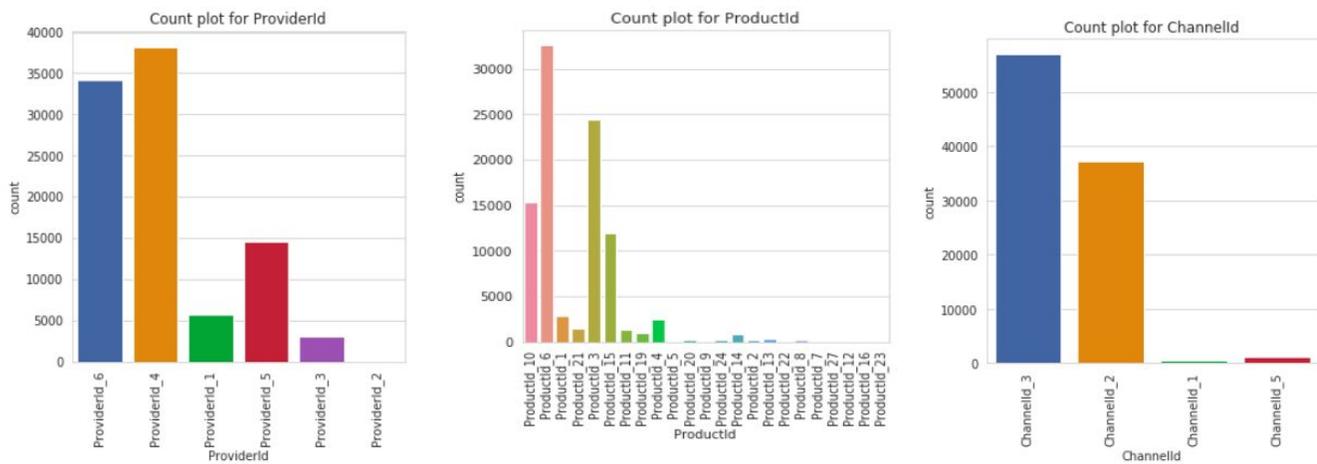

**Fig. 5.** Code to drop redundant columns in a dataset

**Fig. 6.** Output from visualization of categorical features using countplot

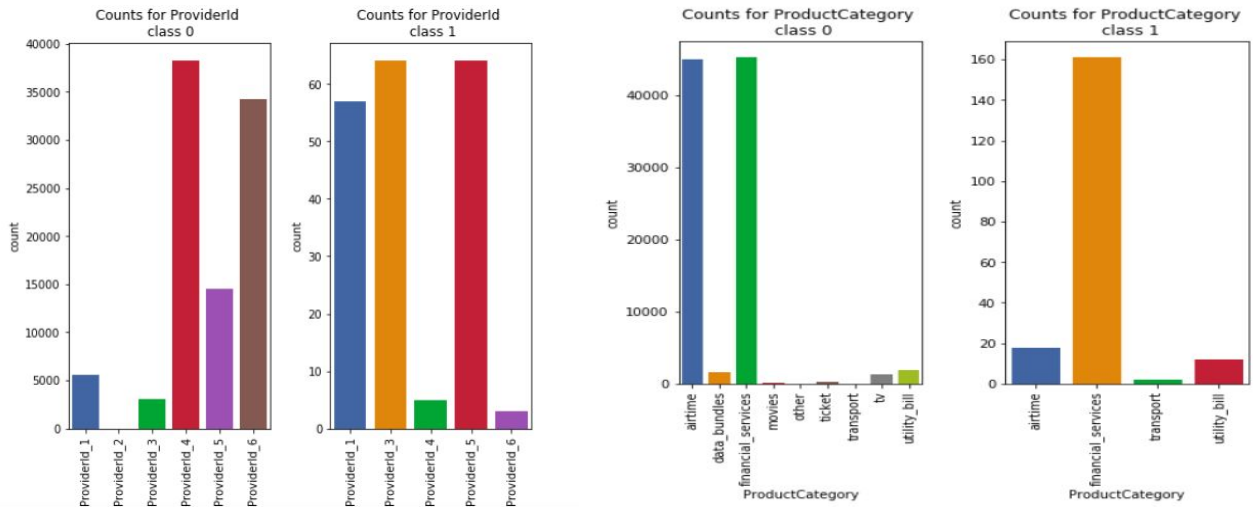

**Fig. 7.** Output from visualization of categorical features using catplot

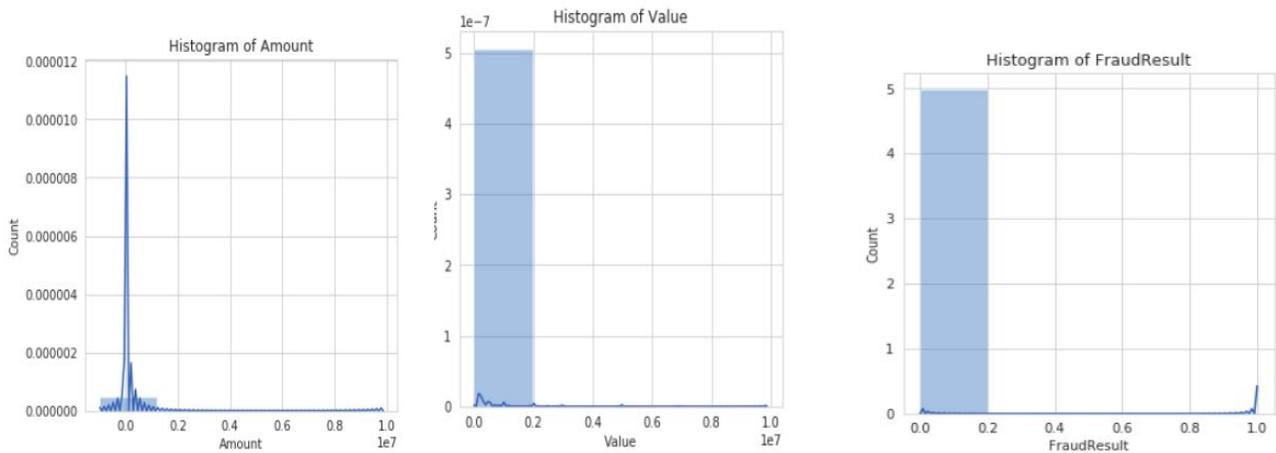

**Fig. 8.** Output from visualization of categorical features using histogram

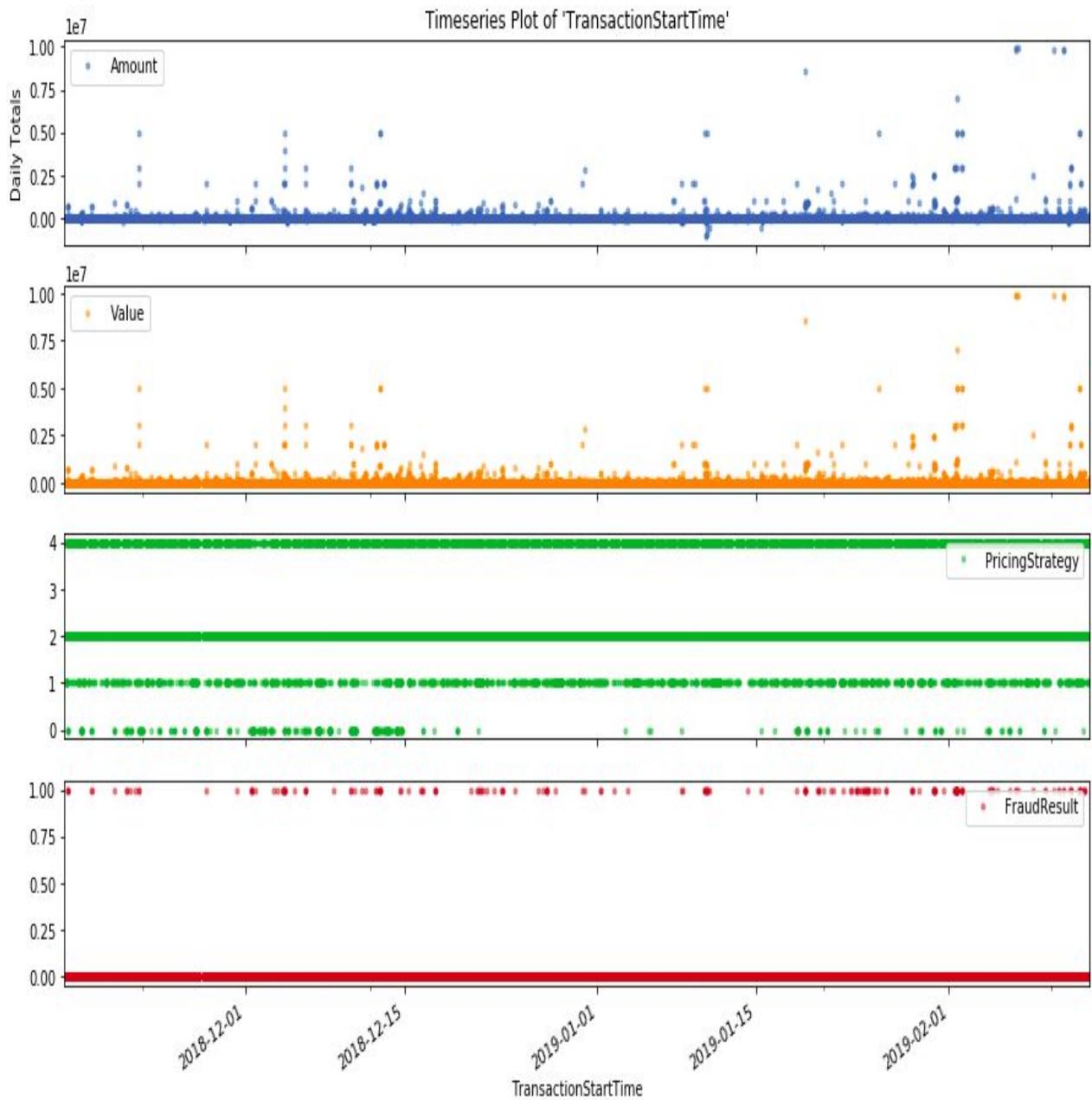

**Fig. 9.** Output from visualization of numerical features against a timestamp feature using the num_timeplot function

```python
train_data = ds.timeseries.extract_dates(data=train_data, date_cols=['TransactionStartTime'])
test_data = ds.timeseries.extract_dates(data=test_data, date_cols=['TransactionStartTime'])
```

|  | 0 | 1 |
|---|---|---|
| AccountId | AccountId_3957 | AccountId_4841 |
| SubscriptionId | SubscriptionId_887 | SubscriptionId_3829 |
| CustomerId | CustomerId_4406 | CustomerId_4406 |
| ProviderId | ProviderId_6 | ProviderId_4 |
| ProductId | ProductId_10 | ProductId_6 |
| ProductCategory | airtime | financial_services |
| ChannelId | ChannelId_3 | ChannelId_2 |
| Amount | 1000 | -20 |
| Value | 1000 | 20 |
| PricingStrategy | 2 | 2 |
| FraudResult | 0 | 0 |
| TransactionStartTime_dow | Thursday | Thursday |
| TransactionStartTime_doy | 319 | 319 |
| TransactionStartTime_dom | 15 | 15 |
| TransactionStartTime_hr | 2 | 2 |
| TransactionStartTime_min | 18 | 19 |
| TransactionStartTime_is_wkd | 0 | 0 |
| TransactionStartTime_yr | 2018 | 2018 |
| TransactionStartTime_qtr | 4 | 4 |
| TransactionStartTime_mth | 11 | 11 |

**Fig. 10.** Extraction of date time features using the extract_dates function

```
from sklearn.linear_model import LogisticRegression
from sklearn.metrics import f1_score
from sklearn.model_selection import cross_val_score, train_test_split

X_train, X_test, y_train, y_test = train_test_split(train, target, test_size=0.3, random_state=2)

rf_model = RandomForestClassifier(n_estimators=100, random_state=232)
lg_model = LogisticRegression(max_iter=100, random_state=2, solver='lbfgs')
```

**Fig. 11.** Importation of models and setting up train and test sets

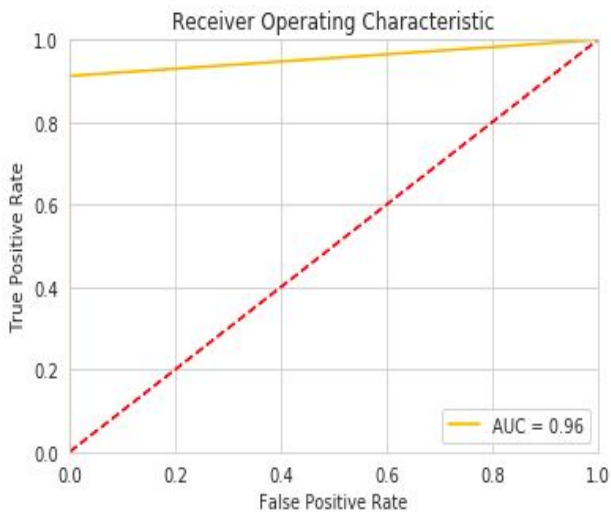
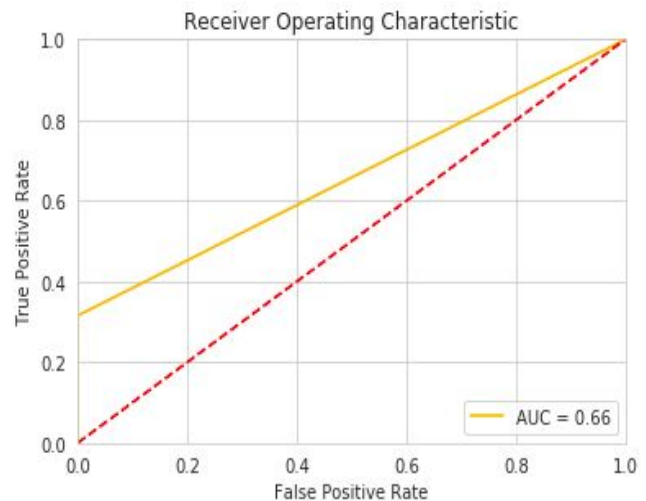

**Fig. 12.** Output from using the get_classification_report function.

```
feats = train.columns
ds.model.plot_feature_importance(estimator=rf_model, col_names=feats)
```

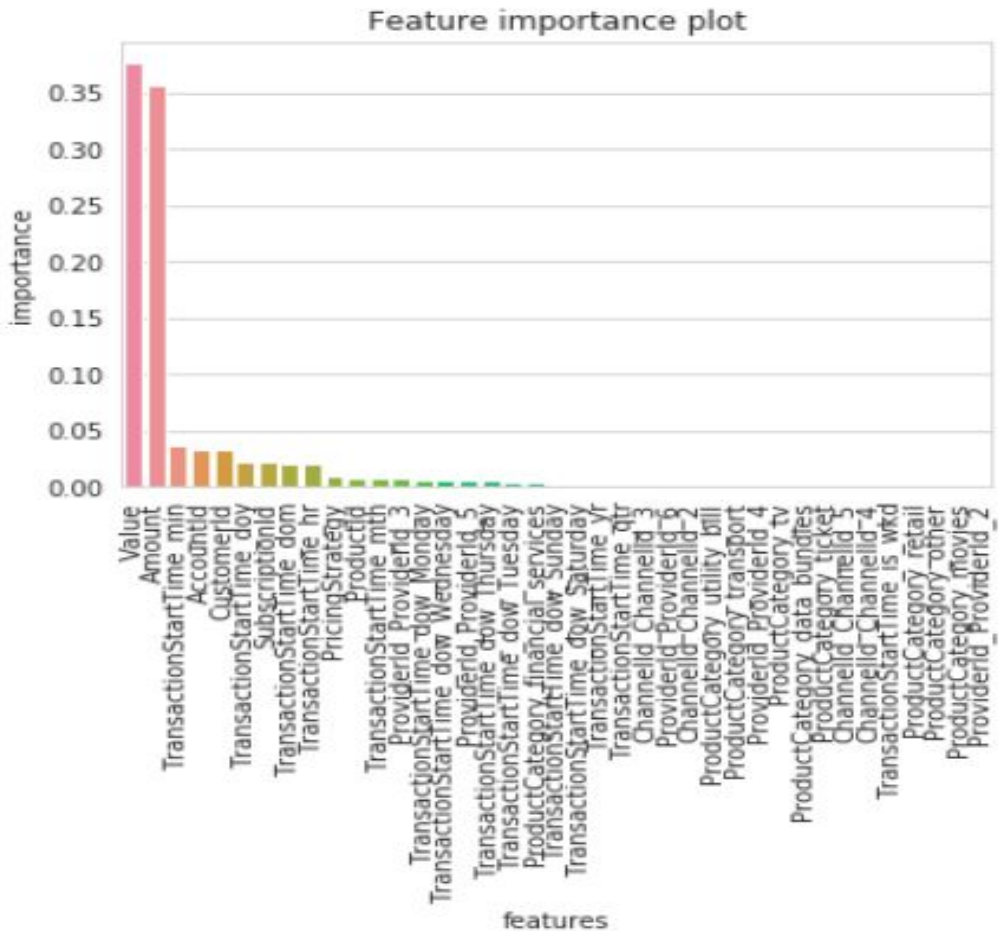

**Fig. 13.** Feature importance plot of the model

## 4      Discussion and Findings

To demonstrate the implementation of our library, we obtained a financial dataset from Xente. The task was posed as a binary classification problem and the feature of interest was the FraudResult feature (Fraud or Not-Fraud). To adequately classify instances as either Fraudulent or Not-Fraudulent, we must clean and prepare the dataset. This preparation was done using our Python library. First we described the data to understand and gain insights from it (see Fig. 4a, 4b and 4c). From this description, we discovered that 5 of the features where numerical, 11 where categorical and 1 was a DateTime feature. Also, we found out that the dataset does not contain missing values, and some of the features where redundant. We removed these features as they were not relevant to the task at hand. To better understand and gain insights from the dataset, we performed visual analysis using visualization module in datasist (see Fig. 6,7,8 and 9). For modeling, we used the LightGBM and RandomForest classifiers. The LightGBM classifiers gave an accuracy of 100%, F1_score of 41%, precision of 58% and recall of 32% and AUC of 0.66 while the RandomForest classifier gave an accuracy of 100%, F1_score of 90%, precision of 90% and recall of 91% and AUC of 0.96 (see Fig. 12). Finally, we plot the feature importance of the classifier see Fig 13), and found out that the features Value and Amount are very important for this predictive task and can greatly help the classifier differentiate fraudulent from non-fraudulent transactions.

## 4.1 Conclusion and Future work

In this research work, we have successfully created a Python based data analytical tool, which aim is to simplify the process of data mining, analysis and visualization. This library can serve as an indispensable tool for data scientists, data analyst and business intelligence experts. We show the basic architecture of our library including the base modules such as feature_engineering, which contains functions and methods relating to feature cleaning and extraction; modeling, which contains functions that help in building and testing machine learning models; structdata, which contains functions for working with structured, tabular datasets; timeseries, which contains temporal related functions and finally visualization module which contains functions for quick visualization of data.

We implemented a possible use case of our library, by mining a fraud dataset provided by a financial company called Xente. The mining task was to create a machine learning model to predict fraudulent transactions. We showed how our library can help in quick data analysis, feature selection and extraction and also how we can use it to extract information from temporal features. Finally, we used a LightGBM and RandomForest classifiers for the modeling task.

In future work, we aim to extend DataSist to support advanced workflows in areas such as Deep learning, Natural language processing, Computer vision, Recommender systems, Data reporting etc.This will provide the same easy off-the-shelf tools and techniques for data scientists working in those areas. Overall, we aim to see DataSist as an integrated work tool in the workflow of data scientists and analysts.

**References**

1. Caldarola, E. G., Picariello, A., and Castelluccia, D. Modern enterprises in the bubble: Why big data matters. ACM SIGSOFT Software Engineering Notes, 40(1):1–4 (2015).
2. Dragland, A. Big data ? for better or worse.˚ ScienceDaily (2013).
3. Franks, B. Taming the big data tidal wave: Finding opportunities in huge data streams with advanced analytics, volume 56. John Wiley & Sons (2012).
4. Earl R. Babbie, The Practice of Social Research, 12th edition, Wadsworth Publishing, ISBN 0-495-59841-0, pp. 436–440 (2009).
5. J. Stirrup, "Tableau dashboard cookbook", 1st ed. Packt Publishing, pp.322. [21] S. Redmond, "QlikView for developers cookbook", 1st ed., 2013, Packt Publishing, pp.272 (2014).
6. Olkin, I., Sampson, A. R., Smelser, Neil J.,Baltes, Paul B., "Multivariate Analysis: Overview", International Encyclopedia of the Social & Behavioral Sciences, Pergamon, pp. 10240–10247, ISBN 9780080430768 (2001).
7. Behrens-Principles and Procedures of Exploratory Data Analysis-American Psychological Association-1997
8. Mohanty, S., Jagadeesh, M., and Srivatsa, H. Big Data Imperatives: Enterprise ?Big Data?Warehouse,?BI?Implementations and Analytics. Apress (2013).
9. Vitaly Friedman. "Data Visualization and Infographics". Graphics (2008).
10. T. Siddiqui and M. Al Kadri, "Big data analytics on the cloud". International Journal of Emerging Technologies in Computational and Applied Sciences (IJETCAS), pp. 61–66 (2015).
11. N. Matloff, "Art of r programming", 1st ed., No Starch Press, Inc., pp.373 (2011).

12. Rud, Olivia. Business Intelligence Success Factors: Tools for Aligning Your Business in the Global Economy, Hoboken, N.J: Wiley & Sons. ISBN (2009) .
13. D. Rotolo and L.Leydesdorff. Matching medline / pubmed data with web of science: a routine in r language", vol. 66, no. 10. In: Journal of the Association for Information Science and Technology, pp. 2155–2159 (2015).
14. Clifton, Christopher. Definition of Data Mining. In: Encyclopædia Britannica (2010).
15. D. Toomey, "R for Data Science", 1st ed., 2014, Packt Publishing, pp.347.
16. "The IPython notebook: a historical retrospective". Fernando Perez Blog. 8 January 2012.
17. C. L. P. Chen and C. Zhang, "Data-intensive applications, challenges, techniques and technologies : a survey on big data", vol. 275, 2014, Information Sciences, pp. 314–347.
18. "Scientific Computing Tools for Python". SciPy.org.
19. Wes McKinney. pandas: a Foundational Python Library for Data Analysis and Statistics (2011).

20. "Python Data Analysis Library – pandas: Python Data Analysis Library". pandas. last accessed 2018/11/13.
21. F. Pedregosa, G. Varoquax, A. Gramfort, V. Michel, B. Thirion, O. Grisel, M. Blondel, P. Prettenhofer, R. Weiss, V. Dubourg, and M. Brucher, "Scikit-learn: machine learning in Python, vol. 12, Journal of Machine Learning Research, pp. 2825–2830 (2011).
22. IBM, "IBM spss statistics 21 brief guide", 1st ed., IBM Corp., pp.158 (2012)..
23. Nie, Norman H; Bent, Dale H; Hadlai Hull, C. SPSS: Statistical package for the social sciences (1970).
24. KDnuggets Annual Software Poll: Analytics/Data mining software used? KDnuggets ( 2013).
25. "Who uses Stata?". Stata. Retrieved 2019/11/15.
26. Hamilton, Lawrence C. Statistics with STATA. Boston: Cengage. ISBN (2013).
27. Zindi, "Data Science Competitions for Africa", https://zindi.africa, last accessed 2019/11/18
28. Xente, "E-commerce, E-payments and E-solution", /https://www.xente.co/, last accessed 2019/11/18